\def\arxivversion{1}
\documentclass{article}
\usepackage{iclr2027_conference}

\usepackage{times}
\usepackage{latexsym}
\usepackage[T1]{fontenc}
\usepackage[utf8]{inputenc}
\usepackage{microtype}
\usepackage{inconsolata}
\usepackage{graphicx}
\usepackage{booktabs}
\usepackage{amsmath}
\usepackage{hyperref}
\usepackage{url}
\hypersetup{hidelinks,pdftitle={Predicting Linear Relational Decodability Without the Relation},
  pdfauthor={},pdfsubject={},pdfkeywords={}}

\title{Predicting Linear Relational\\Decodability Without the Relation}

\author{Anonymous authors}

\ifdefined\arxivversion
  \iclrfinalcopy
\title{\centering Predicting Linear Relational\\Decodability Without the Relation}
\newlength{\arxivauthorsep}
\author{%
Jonas Rohweder\textsuperscript{1,2,3} \And
Adrian M{\"u}lthaler\textsuperscript{1,4} \And
Faridoun Mehri\textsuperscript{1,3} \AND
\makebox[\dimexpr\textwidth-2\tabcolsep\relax][c]{%
Sebastian Gerstner\textsuperscript{1,3}\hspace{\arxivauthorsep}%
Hinrich Sch{\"u}tze\textsuperscript{1,3,4}} \AND
\normalfont
\textsuperscript{1}Center for Information and Language Processing (CIS), LMU Munich\\
\textsuperscript{2}Ubiquitous Knowledge Processing Lab (UKP Lab), TU Darmstadt\\
\textsuperscript{3}Munich Center for Machine Learning (MCML)\\
\textsuperscript{4}relAI – Konrad Zuse School of Excellence in Reliable AI%
}
\hypersetup{pdfauthor={Jonas Rohweder, Adrian Mülthaler, Faridoun Mehri,
  Sebastian Gerstner, Hinrich Schütze}}

\fi

\newcounter{notecounter}
\newcommand{\enotesoff}{\long\gdef\enote##1##2{}}
\newcommand{\enoteson}{\long\gdef\enote##1##2{{
\stepcounter{notecounter}
{\large\bf
\hspace{0cm}\arabic{notecounter} $<<<$ ##1: ##2
$>>>$\hspace{1cm}}}}}

\enoteson
\enotesoff

\def\headlinenumber{0.7355}

\begin{document}
\maketitle
\ifdefined\arxivversion
  \lhead{}
  \renewcommand{\headrulewidth}{0pt}
\fi

\begin{abstract}
Linear relational embeddings (LREs) approximate a language
model's subject-to-object mapping with an affine
function. Their faithfulness varies enormously across
relations, from near zero for \textsc{person\_father} to
the ceiling for \textsc{country\_capital\_city}. We investigate
the factors that contribute to this variation and
demonstrate that it can be significantly predicted from a
readout that is not aware of the underlying relation. To
do so, we propose a novel form of the Jacobian lens, the
\emph{entity lens}. Whereas the Jacobian lens is estimated
across all tokens and concepts, the entity lens
$\mbox{EL}_t$
is trained
to predict what the model will output for an entity of
type $t$ under the neutral prompt "\texttt{\{\} is known for}"; e.g., we train ELs for countries and persons.
\enote{sg}{I don't understand the above sentence: "what the model will output" is too vague. Is it about next-token predictions or about any future token? (Edit after reading the intro: I think we should mention that it's under the neutral prompt "is known for".)}
After estimating an entity lens, we measure how
highly it ranks the model's answer for a relation without
using any relation-specific information in estimation and
prediction.  Across 28 relations of the LRE dataset on
Olmo-3-7B this relation-agnostic score correlates with LRE
faithfulness at $\rho = \headlinenumber$  ($p < .01$).  This
suggests that linear decodability is largely a property of
what a subject's representation already carries, rather than
of the relation being decoded.
Moving the
unit of analysis from the relation to the individual fact,
we conclude that new relation datasets and analysis methods
are needed to address several problems in existing datasets,
most prominently semantically weak first tokens (e.g., most
predicted universities' first token is ``University'') and
reference competition (e.g., the LLM confuses obscure
actor names with competing non-actor entities that have the
same last name).
Our full pipeline is available \href{https://github.com/jonasrohw/entity-lens}{here}.
\end{abstract}

\section{Introduction}

\citet{hernandez2024linearity} showed that a language model's mapping from
a subject to the object of a relation is often well approximated by an
affine function of the subject's hidden state, a \emph{linear relational
embedding} (LRE). They also showed that this works far better for some
relations than others. On OLMo-3-7B \citep{olmo2025olmo3}, we reproduce the same spread: an LRE
fitted on \textsc{country\_capital\_city} agrees with the model's own
prediction 87.5\% of the time while one fitted on \textsc{person\_father}
agrees 7.5\% of the time.

Why? The natural reading is that some relations are ``more linear'' than
others: linearity is a property of the relation. We report evidence
that it is better understood as a property of the \emph{subject's
representation}. Concretely, we show that one can predict how faithful an
LRE will be \emph{before fitting it}, and without using the relation at
all.

Building on recent work by \citet{gurnee2026workspace},
we  introduce the \emph{entity lens} $\mbox{EL}_t$, which
is trained on subjects of type $t$ placed in the neutral prompt
\texttt{"\{\} is known for"} and
evaluates the Jacobian at entities of type $t$. This readout provides factual knowledge about entities of a given type.
The question it answers, \emph{what factual knowledge is this model ready to produce about this specific entity?}, depends on the entity type, but not on the relation.
We estimate this lens at one layer fixed in advance,
and ask how highly it ranks the model's own answer. The lens sees no
relation template, no objects and no (subject,~object) pairs, yet its
score correlates with LRE faithfulness across 28 relations at
$\rho = \headlinenumber $.

We make four contributions. (i)~We introduce the entity
lens, a supervision-free readout of what a model is disposed to say about
an individual entity of an entity type. (ii)~We turn it into a relation-agnostic diagnostic
of linear decodability and show that it predicts LRE faithfulness.
(iii)~We show the prediction holds up
under leave-one-type-out and under controls for object
cardinality and answer-token frequency.
(iv)~Moving
the unit of
analysis from the relation to the individual fact,
we provide an analysis of several
problems in existing datasets that hinder further
progress on understanding LREs in language models, most prominently
semantically
weak first tokens
and reference competition.

\section{Background}
\label{sec:background}

An LRE for relation $r$ approximates the model's computation as
\begin{equation}
  o \;\approx\; \beta\, W s + b_\beta ,
  \label{eq:basiclre}
\end{equation}
where $s$ is the hidden state at the subject's final token and $o$ the
output state. Write $f$ for the model's own map between the two: $f$
sends the subject's hidden state at an intermediate layer $\ell$ to the
final-layer state at the last position. We average local
first-order approximations of $f$ over a \textbf{reference set} $\{s_i\}$ of reference
subjects and scale their subject-dependent terms by $\beta$:
\begin{equation}
W = \operatorname{mean}_i J_i , \qquad
b_\beta = \operatorname{mean}_i\!\left[\, f(s_i) - \beta J_i s_i \,\right] ,
\label{eq:wb}
\end{equation}
where $J_i$ is the Jacobian of $f$ at $s_i$. At $\beta=1$, this is the
averaged first-order Taylor approximation
\citep{hernandez2024linearity}.

This prediction decomposes into a subject-dependent term
and a constant: expanding $b_\beta$ gives
\begin{equation}
\beta W s + b_\beta \;=\;
\operatorname{mean}_i\!\left[\,f(s_i) + \beta J_i (s -
  s_i)\,\right],
\label{eq:beta}
\end{equation}
in which only the second term depends on which subject $s$ is being decoded.
Left unscaled, the shared anchor $\operatorname{mean}_i f(s_i)$ dominates
and the operator returns much the same object for every subject.
Setting $\beta > 1$ amplifies the subject-specific contribution relative
to that anchor, and so does change which object is ranked first.
\citet{hernandez2024linearity} fit $\beta$ per relation alongside the
layer; we follow their protocol and select it from
$\{0.5, 1, 1.5, 2, 3, 5\}$.

\emph{Faithfulness} is the rate at which
decoding $\beta W s + b_\beta$ over the full vocabulary reproduces the
\emph{model's own} top-1 prediction (not the gold object). It is therefore
a measure of how well an affine map captures what the model
does. Faithfulness defined this way is
the quantity we predict.

\section{Related work}

\textbf{Linear relational embeddings.}
\citet{DBLP:conf/icml/PaccanaroH00} proposed Linear
Relational Embeddings (LREs) to model
relations as linear maps
between concept representations.
\citet{hernandez2024linearity} applied LREs
to transformers \citep{vaswani2017attention}.
Subsequent work has asked why LREs are faithful for some
relations and not others.
\citet{merullo2025linear} trace part of this
variation to pretraining data: LREs are more faithful when subjects and
objects frequently co-occur in the corpus. Co-occurrence frequency also predicts whether
facts are acquired during pretraining, though it does not fully explain
learning \citep{10.1162/TACL.a.746}. We report such corpus-frequency measures as baselines in Appendix~\ref{sec:corpus-freq-baselines}. A further question is whether such
structure reflects what a model has learned or only the particular
parametrization that realizes it, since many parametrizations induce the
same next-token distribution. Existing results point to the former,
since under suitable conditions, relational linearity holds either in
all distribution-equivalent models or in none \citep{marconato2025all}.
Using a KL-divergence probe,
\citet{valer2026relational} have shown
that relational linearity varies across models, layers and phrasings of the relation.
These works evaluate and explain linearity at the level of the relation.
We ask instead how much of it can be predicted from how the model
represents the subject, using a readout estimated without the relation
or any of its objects.

\textbf{Subject representations in factual recall.} Mechanistic studies
converge on the view that much factual knowledge is attached to the
subject before the relation is processed.
Feed-forward layers operate as key-value memories whose keys respond
to patterns in the input and whose values induce distributions over the
output vocabulary \citep{geva2021transformer}.
\citet{meng2022locating} show that it is in the middle
layers at the last subject token that
factual
associations are localized.
Building on these findings, factual recall has been decomposed into
stages: MLP layers enrich the subject representation with many of its
attributes, a lookup that depends largely on the subject's name rather
than on the preceding context; the relation is propagated to the final
position; and attention heads extract the requested attribute
\citep{geva-etal-2023-dissecting, nanda2024fact}. Our approach can be read as a probe of this
enrichment, since it reports what the model is prepared to say about a
subject before any relation is given. This allows us to connect two
lines of work: We hypothesize that the enrichment the lens diagnoses
contributes to the relational linearity found by
\citet{hernandez2024linearity}. The lens predicts LRE
faithfulness across relations in all seven models we test, and the association persists when controlling for how many subjects
share each object and for answer-token frequency (§\ref{sec:results}).

\textbf{How facts are stored.} Theoretical work has analyzed two mechanisms by which transformers store facts.
\citet{nichani2025understanding} model attention and MLP
weights as associative memories and prove that a one-layer transformer
can store a number of facts that scales linearly with its parameter
count, with storage traded off between self-attention and MLP
parameters. \citet{DBLP:journals/corr/abs-2605-12426} study a
geometric form of storage: in a controlled one-layer setting, subject
embeddings encode linear superpositions of their associated attributes,
and a small MLP selects among them conditioned on the relation. In
pretrained models the two need not compete: lookup-like computation can
enrich the subject representation over the early and middle layers, and
the geometric picture then describes the enriched state from which a
relation selects. Our hypothesis concerns this enriched state: a
relation should be linearly decodable to the extent that its attribute
is prominently encoded in the subject representation after enrichment.
\enote{hs}{nice!}

\textbf{Lenses.} Lenses read out intermediate hidden states in terms of the output
vocabulary and differ mainly in the map they apply: the model's own
unembedding \citep{nostalgebraist2020interpreting} or an affine
translator learned per layer \citep{belrose2023eliciting}. Patchscopes
place these readouts in a common framework, in which a hidden state is
decoded by patching it into a target prompt, and have been used to
inspect how models resolve entity names
\citep{ghandeharioun2024patchscopes}. More recently,
\citet{gurnee2026workspace} apply the model's input-output Jacobian,
averaged over a large corpus, to raw activations, with the aim of
identifying representations the model is poised to verbalize. Our entity lens instead evaluates the Jacobian at reference entities of the subject's type under a single neutral factual prompt, so that the
readout concerns what the model is prepared to say about entities of
that type rather than about any concept. Its contrastive variant further
applies the Jacobian to differences between entity representations,
subtracting what the references share.

%
%
%
%
%

\section{The entity lens}
\label{sec:lens}
\enote{sg}{I think the whole CEL stuff is something we could move to the appendix, so the reader can focus on NEL which now seems to be the main thing.}
\enote{sg}{Regarding my above comment: I was told this would break the logic of this section. But actually, there's a consistent single-line explanation for going from LRE to NEL: We drop all terms that do not depend on $s$.}

\citet{gurnee2026workspace}'s Jacobian lens is broadly
similar to LRE.  But
instead of a relation-specific prompt and a focused target
position (the object),
it computes the average 
$W = \operatorname{mean}_i J_i$ over many diverse prompts and many
output positions. The Jacobian lens also does not use the
``anchor'', the zero-order term of the Taylor expansion; so it
is simply $W = \operatorname{mean}_i J_i$.

We derive our \emph{contrastive entity lens} from
Eq.~\ref{eq:beta}, restated as:
\begin{equation}
  f(s) \approx
\beta W s + b_\beta \;=\;
\operatorname{mean}_i\!\left[\,f(s_i)\right]
+ \beta\operatorname{mean}_i\!\left[\, J_i (s - s_i)\,\right]
\end{equation}
Following \citet{gurnee2026workspace}, we also drop the anchor
term $\operatorname{mean}_i\!\left[\,f(s_i)\right]$; see discussion in 
\S\ref{sec:swap}.
This also means that we do not need $\beta$ since the
scaling does not change the output distribution over
tokens. The final form of the contrastive entity lens is then:
\begin{equation}
\mbox{CEL}(e) = \operatorname{mean}_i\!\left[\, J_i (s_e - s_i)\,\right]
\label{eq:cel}
\end{equation}
CEL takes into account the difference between $s_e$ and
the reference point $s_i$, as it should if we want
to approximate $f(s_e)$ using the first derivative. 
Our original motivation was that we are not interested in what the model associates
with an entity like ``Russia'' in general (e.g.,
``Europe''). Instead we are interested in what the model
associates with ``Russia'', but not with other reference
entities like ``Spain'' and ``Italy''. Examples of such
contrastive associations for ``Russia'' are ``vodka'' and ``Kremlin''.
Our hypothesis was that such top-ranked contrastive associations
give rise to linearity.

In initial experiments, we found that the contrastive entity
lens is quite sensitive to the reference set. For example,
if $s_e$ is a French person and a Spanish person is the
reference entity $s_i$, then the vector difference in 
Eq.~\ref{eq:cel} will enhance ``French''. But if $s_i$ is
another French person, then ``French'' will be demoted.
This is bad for answering questions about $s_e$'s
nationality and native language. We therefore also consider
the following noncontrastive entity lens, which drops the
reference set differences:
\begin{equation}
\mbox{NEL}(e) = \operatorname{mean}_i\!\left[\, J_i s_e \,\right]
\label{eq:nel}
\end{equation}
In what follows,
EL refers to both NEL and CEL.
\enote{sg}{should we refer to \S\ref{sec:refsets} for more info about the CEL-NEL contrast?}


To instantiate EL for factual
relations, we need (i) a prompt that evokes factual associations
and (ii) a reference set of entities for computing the Jacobians $J_i$
(both NEL and CEL) and for serving as the $s_i$ in the
vector differences $s_e - s_i$ (for CEL).

As the prompt we adopt \texttt{"\{\} is known for"}. This is a neutral prompt, i.e.,
it is not customized for any specific factual relation. We
use it for all our  experiments unless otherwise stated.

As the reference set, we use entities of the same type as
the subject, e.g., other countries if the subject is a
country, other landmarks if the subject is a landmark
etc.
NEL applied to an input $s$ outputs anything the model is likely
to say in a context in which an entity of the same type as $s$ occurs, including features
it shares with other same-type entities.
CEL outputs only what differentiates
$s$ from other same-type entities.

\enote{sg}{Idea for a future iteration: what if we train just one NEL for all subject types together?}

\paragraph{Computation.} We do not compute $J_i$ explicitly, but obtain each
Jacobian--vector product $J_i u$ by the double-backward
construction \citep{pearlmutter1994fast}, at the cost of a constant
number of forward and backward passes (Appendix~\ref{sec:jvp}).


\enote{hs}{is this clearly stated in this section? or do we
  need to repeat it here?
\paragraph{What makes $A_r$ relation-agnostic.} Eq.~\ref{eq:cel} is not specific to $r$: the prompt is neutral and shared,
the layer is fixed at 17 for all relations
(Appendix~\ref{sec:layer17}) and
no (subject,~object) pair of the
relation is used in estimation.
}

\enote{sg}{I think we need to briefly mention layer 17 somewhere. (Btw. it is probably not the same layer for all models?)
The rest is clear enough in my opinion.
}

\section{Experimental setup}
\label{sec:setup}

\paragraph{Models.} Our primary analysis is conducted on OLMo-3-7B because it combines performance with full replication resources for analysis. Results for six additional models appear in Appendix~\ref{sec:shared-baselines}. Inference details are in Appendix~\ref{sec:jvp}.

\paragraph{Relations.} The LRE dataset \citep{hernandez2024linearity}
contains 47 relations. Our prompt presupposes an entity subject, so we
fixed a subject-type filter in advance using the dataset's declared
\texttt{domain\_name} field: the 28 relations with named-entity subjects
(\textit{person}, \textit{country}, \textit{name}, \textit{company},
\textit{landmark}, \textit{superhero}, \textit{city}, \textit{pokemon},
\textit{product}, \textit{star}). Excluded are relations over words as
words (\textit{verb}, \textit{adjective}, \textit{word}) and over abstracts
(\textit{occupation}, \textit{task}, \textit{characteristic},
\textit{degree}), for which \texttt{"\{\} is known for"} is a category
error. No other relation was excluded.


\paragraph{Measurement.}
For LRE faithfulness,
we follow
\citet{hernandez2024linearity}. A relation is presented to the model
\emph{in context}: eight (subject, object) examples of the relation are
concatenated, one per line, and the query subject is appended without its
object, so that the model's next token is its answer.
This is what
defines $m_e$, the model's \textbf{top-1 answer}. The subjects of
a relation are shuffled once and partitioned into three disjoint sets:
eight in-context examples and eight reference subjects from which $W$ and $b_\beta$
are estimated.

Faithfulness is then recomputed for all 28 relations under this protocol.
Layer and $\beta$ are chosen on a $6\times6$ grid consisting of six candidate
layers for the subject state ($5, 11, 14, 17, 20, 23$) crossed with six values
of $\beta$ ($0.5, 1, 1.5, 2, 3, 5$). Selection is based on the mean over the
test subjects.
\citet{hernandez2024linearity} likewise fit both per relation; the six
layers span the network at roughly even spacing and the $\beta$ range
brackets the values they report.

To measure the accuracy of the predictions of the entity lens,
we decode $\mbox{EL}(e)$ over the full vocabulary
and record the rank of $m_e$, the model's top-1 answer.
Writing $\mathrm{rk}_{\mbox{EL}(e)}(m_e)$ for that rank, the diagnostic for a
relation is
\begin{equation}
A_r \;=\; -\operatorname{median}_{e}\,\log_{10}\mathrm{rk}_{\mbox{EL}(e)}(m_e),
\end{equation}
so higher $A_r$ means the lens agrees with the model more.

The lens and the LRE score every row of a relation (19--994 rows), including
the in-context examples and the reference set. For the
results in \S\ref{sec:results}, the lens and the LRE sample their splits
independently, so the particular subjects entering each set differ between
the two measurements. Independent splits introduce sampling variation into
this comparison. Because the LRE hyperparameters are selected on the scored
subjects, its reported faithfulness is an in-sample estimate; these results
do not establish performance on a separate held-out evaluation set.

Note that the in-context prompt serves only to define the target
$m_e$; the lens itself is always
computed under the neutral prompt of \S\ref{sec:lens}.

\section{Results}
\label{sec:results}

\begin{figure}[t]
\centering
\includegraphics[width=0.8\linewidth]{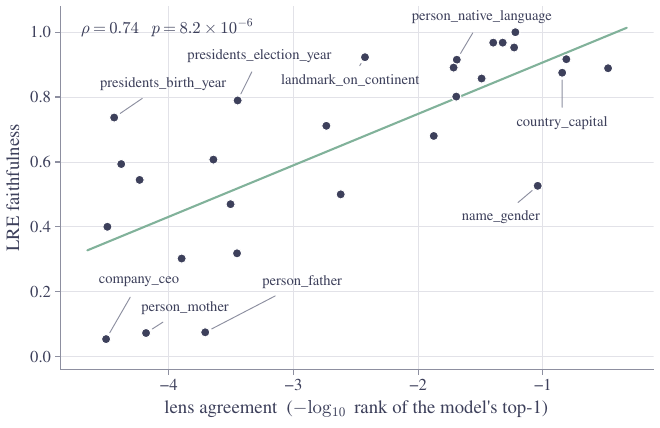}
\caption{LRE faithfulness against the relation-agnostic CEL lens score, 28
relations. The lens sees no relation template, no objects, and no
(subject, object) pairs, and uses one fixed layer throughout.}
\label{fig:main}
\end{figure}

Figure~\ref{fig:main} plots LRE faithfulness against the median entity-lens
score across subjects for each of 28 relations, with a least-squares fit.
The scores correlate at Spearman $\rho = \headlinenumber$
($p = 8.24\!\times\!10^{-6}$).
We next test three other predictors of faithfulness.


\enote{hs}{pleaes move this footnote to the appendix
footnote{All variables here are ordinal, so we work throughout with
ranks. To remove the effects of a set of covariates, we
rank-transform the lens score, faithfulness, and each
covariate; regress the ranked lens score and ranked
faithfulness separately on the ranked covariates using
ordinary least squares with an intercept; and report the
Pearson correlation between the resulting residual vectors.
This is the standard partial
Spearman correlation, and it asks how much of the lens--faithfulness
association remains once everything the covariates can linearly explain
about either variable has been removed.}
}

\enote{sg}{I moved the object cardinality paragraph here because the next paragraph (answer frequency) assumes the reader knows what "cardinality" is.}
\textbf{Object cardinality.} Relations
differ greatly in how many distinct objects they have; $k$, the mean number
of subjects per object, ranges from $1.0$ to $189$. High $k$ should make
faithfulness easier, and $k$ is indeed associated with faithfulness on
its own, though only weakly
($\rho = 0.355$, $p = 0.064$). But $k$ is weakly correlated with the
lens score ($\rho = 0.086$). Removing $k$  \emph{raises} the
correlation to $\rho = 0.757$ ($p = 4.90\!\times\!10^{-6}$): $k$ and the lens are
complementary predictors of faithfulness, so removing the variance
attributable to $k$ sharpens rather than weakens the lens
signal. Since object cardinality is a complementary
predictor, we will analyze it in more detail in \S\ref{sec:residuals} below.

\textbf{Answer frequency.} A relation whose answers are frequent
tokens might be easy for both measures. We estimate the model's prior over
each answer token by averaging its log probability over ten short
generic prefixes.
This predicts neither faithfulness ($\rho = -0.111$) nor the lens
score ($\rho = 0.098$), and removing it leaves the correlation
unchanged at $0.755$. The correlation is $0.784$ with both cardinality
and answer frequency removed.\enote{sg}{We need to explain "cardinality" the first time we use it.}

\enote{hs}{
  please move to appendix
footnote{The beginning-of-sequence token alone,
``The'', ``In the'', ``It was'', ``She said the'', ``Here is'', ``One of
the'', ``According to the'', ``There is a'', and ``Yesterday the''.}
}

\textbf{Entity type.} Eleven of the 28 relations have
\textit{person} subjects, raising the worry that one domain
(i.e., one subject type)
drives the
result. Holding out each subject domain in turn leaves
$\rho \in [0.696, 0.799]$, median $0.728$.

\subsection{Object cardinality}
\label{sec:residuals}
It turns out that object cardinality partially explains the
tails of the distribution 
in Figure~\ref{fig:main}.
We now investigate  the two tails, i.e., those relations
with the largest positive and negative deviations
in Figure~\ref{fig:main}.

We rank relations by their residuals from the least-squares fit in
Figure~\ref{fig:main}, excluding the two rescored year relations
(Appendix~\ref{sec:numeric-objects}). For \textsc{name\_gender}, we use
the model's literal answer token, matching faithfulness, rather than all
gender-matched alternatives. A residual of $+0.30$ means faithfulness
is thirty percentage points above the lens prediction.


The largest positive residuals, faithful despite a low
lens score, with their cardinality are
\textsc{name\_gender} ($+0.289$, $k = 9.5$),
\textsc{person\_university} ($+0.258$) and
\textsc{landmark\_on\_continent} ($+0.239$, $k = 189$). In each, the LRE
chooses among a handful of possible answers (for \textsc{person\_university},
the model answers ``University'' for 54 of 91 subjects; \S\ref{sec:subject}), and can be faithful without
the answer being salient for any particular subject: no name is
chiefly ``known for'' its gender, but there are only two answers to pick
between.

The largest negative residuals, unfaithful despite the
lens, are \textsc{person\_father} ($-0.381$),
\textsc{person\_mother} ($-0.299$) and \textsc{company\_ceo}
($-0.260$), all with $k \approx 1$ and answer spaces in the
hundreds. For these relations, faithfulness is near its
floor due to the difficulty of the task and the details of
the representation apparently do not change that.

The relationship between our NEL-based score and LRE
faithfulness thus holds in the middle of the cardinality
range and degrades at both ends, in opposite directions;
across relations, residuals increase with $k$ (Spearman
$\rho = 0.60$, $p = 0.001$).
Our analysis in this subsection helps us interpret the
result from the beginning of this section: $k$ is a true
source of deviation, not merely a confound.

\enote{sg}{In general I think we should avoid using the word "account" too much, it's a sign of AI slang}

\begin{table}[t]
\centering\small
\caption{Object prominence, measured by the model's own free
generation and therefore independently of the lens, against how often
the lens surfaces the gold object in its top five. Spearman
$\rho = 0.570$ ($p = 0.033$, $n = 14$).}
\label{tab:prominence}

\begin{tabular}{@{}lrrrr@{}}
\toprule
relation & prom.  & gold$\in$lens & $n$ \\
\midrule
city\_in\_country & 0.96  & 0.71 & 24 \\
person\_plays\_pro\_sport & 0.69  & 0.36 & 318 \\
landmark\_in\_country & 0.56  & 0.30 & 836 \\
star\_constellation & 0.48  & 0.06 & 361 \\
person\_university & 0.32  & 0.00 & 91 \\
country\_capital\_city & 0.29  & 0.42 & 24 \\
person\_occupation & 0.25  & 0.12 & 821 \\
company\_hq & 0.23  & 0.02 & 674 \\
product\_by\_company & 0.22  & 0.19 & 522 \\
country\_largest\_city & 0.21  & 0.42 & 24 \\
country\_currency & 0.20 & 0.03 & 30 \\
person\_father & 0.11 & 0.01 & 942 \\
person\_mother & 0.10  & 0.01 & 986 \\
superhero\_archnemesis & 0.06  & 0.01 & 94 \\
\bottomrule
\end{tabular}
\end{table}

\section{Analysis of the entity lens}
\subsection{Salience}
\label{sec:salience}
Our approach is based on an assumption that is also shared
by the Jacobian lens:
Readouts of this kind
are normally taken to surface what a
model is internally disposed to say
\citep{gurnee2026workspace}. We read $\mbox{EL}(s)$ the same way: as what the subject's representation makes
available, before any relation has been named. Our interpretation of
Figure~\ref{fig:main} rests on that reading. If it is right, a relation
should be linearly decodable to the extent that its answer is already
part of what the subject evokes; that is, the object should be
\emph{salient} for the subject.
This section performs a sanity check
(a check that is external to the lens)
on that assumption by
measuring salience by the
model's own free generation.

For each test subject we greedily generate 90 tokens
from each of \texttt{"\{\} is best known for"} and \texttt{"Here are some
facts about \{\}. First,"}, and record whether any distinctive word of the
gold object appears in either continuation.


\enote{hs}{please move this footnote to the appendix
footnote{Titles, particles and
words shared with the subject's own name are excluded, so that
\emph{The}~Joker is not matched by any occurrence of ``the'' and
\emph{Johann Georg Kant} (Immanuel Kant's father) is not matched by a passage about Immanuel Kant.
The lens's own prompt cannot be used here: \texttt{"\{\} is known for"}
makes Olmo-3-7B emit a multiple-choice quiz rather than prose, so the
object is never named.}
}


We refer to this rate as the object's \emph{prominence}.
Table~\ref{tab:prominence} gives both prominence and
gold$\in$lens
(how often the lens surfaces the
gold object in its top five)
for 14 relations.
Prominence
ranges from $0.96$
(\textsc{city\_in\_country}) and $0.29$ (\textsc{country\_capital\_city})
down to $0.11$ (\textsc{person\_father}), $0.10$
(\textsc{person\_mother}) and $0.06$
(\textsc{superhero\_archnemesis}). It predicts how often the lens surfaces the
gold object in its top five: Spearman $\rho = 0.570$ ($p = 0.033$,
$n = 14$).
Thus, what the lens surfaces is predicted by a 
salience measurement that is distinct from the salience
measured by the lens.


\subsection{Reference sets}
\label{sec:refsets}
\enote{sg}{We should move this section to the appendix. It is only relevant for the CEL-NEL contrast, and it is way too detailed and AI-y.}

\S\ref{sec:lens} motivated NEL by the observation that CEL is sensitive
to the reference set: if the reference entities $s_i$ share the answer
with $s_e$, the difference $s_e - s_i$ removes it. We now
measure that effect.

We test \textsc{person\_native\_language} on a fixed subset of 100 subjects
(Table~\ref{tab:refsets}): 69 are French speakers, and the model answers
``French'' for 65 of them, allowing us to observe cancellation of a
single-token answer. We preselect eight French and eight non-French
reference entities and exclude them from testing. All four reference sets
are evaluated on the same 61 French and 23 non-French subjects.

\begin{table}[t]
\centering\small
\caption{The reference set's effect on the two lenses, on
\textsc{person\_native\_language}. All four reference sets are scored on
the same 61 French subjects. Left: median $\log_{10}$ rank of the
model's own answer $m_e$. Right: median rank of the token ``French''
itself. CEL degrades as the reference entities come to share the
subject's answer; NEL, which takes no difference, does not.}
\label{tab:refsets}

\begin{tabular}{@{}lrrrr@{}}
\toprule
& \multicolumn{2}{c}{median $\log_{10}$ rank of $m_e$}
& \multicolumn{2}{c}{median rank of ``French''} \\
\cmidrule(lr){2-3}\cmidrule(lr){4-5}
reference set & CEL & NEL & CEL & NEL \\
\midrule
8 non-French & 1.62 & 1.81 & 47 & 64 \\
8 from other relations & 1.88 & 1.72 & 102 & 70 \\
4 French, 4 non-French & 1.93 & 1.79 & 115 & 82 \\
8 French & 2.72 & 1.71 & 665 & 55 \\
\bottomrule
\end{tabular}
\end{table}

Table~\ref{tab:refsets} shows the result. Under CEL, moving from
non-French to French reference entities costs $1.10$ in median
$\log_{10}$ rank, a factor of 13.
In contrast, NEL is robust to reference set composition:
results are in a small range, 
$[1.71,1.81]$. Tracking the answer token
itself rather than each subject's own target makes the mechanism plain:
the rank of ``French'' rises monotonically with the number of French
reference entities, $47 \to 115 \to 665$ for zero, four and eight, and
the effect is monotone for 52 of the 61 subjects individually. The
control is the last column. \emph{The same eight French reference
entities} leave ``French'' at rank $55$ under NEL, its best value for
any reference set. The collapse is therefore a property of the subtraction,
not of the reference entities.

\enote{hs}{i think we can cut this?
paragraph{An example.} For Joseph Laniel, whose answer is ``French'',
NEL ranks ``French'' first. CEL ranks it third with non-French reference
entities, and $623$rd with French ones, where the top of the readout is
``sculpt'', ``dancing'', ``singing'', ``dance'', ``chore'', ``piano''.
The subject, the layer and the prompt are identical in the two cases;
only the reference set differs.}

\textbf{Reversal of the effect.} On the 23 non-French subjects the
ordering under CEL is the other way round: the eight French reference entities are
the \emph{best} set (median $\log_{10}$ rank $0.78$) and the eight
non-French ones the worst ($1.97$). The same reference set helps or
hurts depending on what is being asked, so reference quality is not a
property of the set alone.

\textbf{What is subtracted.} CEL is
$\operatorname{mean}_i[J_i s_e] - \operatorname{mean}_i[J_i s_i]$, and
it is tempting to read the second term as the answer. It is not: decoded
over the vocabulary it gives ``Cin'', ``La'', ``Une'', ``fils''. These are
French vocabulary items, not the token ``French''. This is expected, since
$\operatorname{mean}_i[J_i s_i]$ is what the model has loaded about those
people under \texttt{"\{\} is known for"}, which is French cultural
material rather than an answer to a question about language. The answer
is suppressed because the two terms share the French direction, not
because the subtracted term is the answer.

\textbf{Summary.} A reference set of the same type as the subject
(\S\ref{sec:lens}) does not guarantee that the set is neutral with
respect to the relation being read out, and no choice of reference
entities can be neutral for every relation at once. This is why
NEL is preferred to CEL.

\subsection{An interventional analysis}
\label{sec:swap}
\enote{sg}{I think this section is another good candidate for the appendix.}

The LRE of Eq.~\ref{eq:beta} is the sum of an \emph{anchor}
$b_\beta$
and a \emph{subject term}
$\beta Ws$. The subject term has the form of NEL
(Eq.~\ref{eq:nel}).
\enote{sg}{Doesn't the subject term have the form of CEL? (Btw "has the form of" seems like another AI-ism.)}
We recombine the terms to determine what each captures. If the anchor captures the relation, pairing it with a
\emph{foreign} subject term from another relation should still
produce this relation's object;
and if NEL \enote{sg}{or CEL? (see previous enote)} captures the subject, it
should be able to stand in for the subject term.
\enote{sg}{The contrast "if the anchor captures the relation" vs. "if NEL/CEL captures the subject" doesn't seem clean: shouldn't we vary one parameter at a time?}


We use two relation pairs that share their subjects:
\textsc{country\_capital\_city} and \textsc{country\_largest\_city} over
22 countries whose capital is not their largest city and for which the
model answers both relations correctly, and
\textsc{person\_plays\_position\_in\_sport} and
\textsc{person\_plays\_pro\_sport} over 24 athletes. Each subject is read
with the others as references, at $\beta = 3$.
Appendix~\ref{sec:swapdetail} describes the experiments in full.

\begin{table}[t]
\centering
\small
\begin{tabular}{lrrrrrr}
\toprule
& & \multicolumn{3}{c}{subject term} & \multicolumn{2}{c}{controls} \\
\cmidrule(lr){3-5}\cmidrule(lr){6-7}
relation & $n$ & own & foreign & NEL & anchor alone & wrong donor \\
\midrule
\textsc{country\_capital\_city} & 22 & 12 & 12 & 7 & 0 & 0 \\
\textsc{country\_largest\_city} & 22 & 12 & 8 & 4 & 0 & 0 \\
\textsc{person\_plays\_position} & 24 & 19 & 21 & 18 & 7 & 6 \\
\textsc{person\_plays\_pro\_sport} & 24 & 22 & 21 & 21 & 6 & 3 \\
\bottomrule
\end{tabular}
\caption{Recombining an LRE's anchor with different subject terms,
$\beta = 3$. Counts are how often the decoded top-1 is the relation's
gold object. \emph{Own} is the unmodified operator; \emph{foreign} uses
the paired relation's subject term; \emph{NEL} is the
(relation-agnostic)
noncontrastive
entity lens; \emph{anchor alone} drops the
subject term; \emph{wrong donor} uses another subject type's NEL. The model
answers all 22 countries correctly, but the operator itself reproduces
only 12 (Appendix~\ref{sec:swapdetail}).}
\label{tab:swap}
\end{table}

\paragraph{The anchor captures the relation.} A foreign subject term
leaves the answer largely intact (Table~\ref{tab:swap}): with the
largest-city subject term, the capital anchor still names 12 capitals, as
many as with its own; each person anchor keeps 21 of 24 subjects with
the other relation's subject term. The anchor alone names no country's
capital or largest city, so these hits are composed rather than read off
the anchor. Nor do they depend on the particular references: over 500
anchors estimated from references disjoint from the test subjects, the
foreign subject term stays on average within one to two subjects of the
operator's own on all four relations.

\paragraph{The lens stands in for the subject term, but only
moderately.} NEL, computed under the neutral prompt without
relation-specific information,
recovers the operator for sport (21 against 22) and largely for
position (18 against 19), but for only 7 of the 12 capitals and 4 of the
12 largest cities that the operator itself gets right. What it supplies
is subject-specific: another subject's NEL falls to or below the
anchor alone. It is also less stable than the foreign subject
term. Over the same 500 anchors it ranges from 7 to 20 of 24 for
position, and for the country pair neither the anchor's references nor
NEL's own references dominate its variance.

\section{Fact-level analysis}
\label{sec:subject}
Linearity emerges as a phenomenon at the relation level, not
the fact level. It would not make sense to investigate
linearity for the facts of a relation that is not
represented linearly. But
once there is evidence for at least partial linearity of a
relation, we determine for individual facts how well
the entity lens predicts the degree to which
they are
integrated into the linear structure.


To analyze relations at the fact level, we compute 
$\rho$ between lens $\log_{10}$ rank and LRE $\log_{10}$
rank for each relation.
The values of $\rho$ are in the range [-0.401,0.664] with a
median of
0.235. 
We found several shortcomings in the experimental setup currently used to investigate relations in LMs that result in these low values.
We now analyze these shortcomings and give
guidance for future relation datasets that may better support
research on how LMs represent facts.


\subsection{Reference competition}
\label{sec:competition}
The three relations 
\textsc{person\_university} ($\rho$=-0.401),
\textsc{person\_occupation} ($\rho$=-0.388) and
\textsc{person\_plays\_instrument} ($\rho$=-0.035) have the
lowest correlations.
They
exemplify the problem of 
\emph{reference competition}. An example is Rolf Thommessen
who is listed as an actor
in LRE. The highest ranked token given by the lens is
``whistlebl''. This indicates that the most prominent entity
associated with the last name Thommessen
is Advokatfirmaet
Thommessen, a law firm known for digital whistleblowing.
Apparently,  the first name ``Rolf''
does not disambiguate which subject is intended.
In general, the last few tokens seem to
dominate which entity is represented and used by the LM. For the actor Barry Sobel, the lens ranks
``Jewish'', ``Israel'' and ``Holocaust'' among its top five, associations for
the Brazilian Rabbi Henry Sobel. 
The relation \textsc{star\_constellation}
(which in Table~\ref{tab:prominence} was an outlier: high
prominence, low gold$\in$lens)
also suffers from
reference competition due to the complex names of
many stars. For example, the most prominent associations of
the star
``rho01 Cnc''
are ``machining'', ``CNC'', ``manufacturing''
(Cnc = Computerized Numerical Control).

\enote{hs}{cut for space
For the actress Vondell
Darr, the lens top-ranks ``criminal'', ``murders'',
``felon'', plausible associations for the wrongfully convicted
Acacia Lyn-Darr.  For the actor Simon Gosejohann, the lens
top-ranks ``Johann'', ``Hann'', ``German'', suggesting the LM
simply uses the string ``Johann'' as the representation.
While we did not conduct an extensive analysis, it is likely
that reference competition is a problem for many relations whose
subjects are persons. Many of those persons share the
properties that result in Barry Sobel, Vondell Darr
and Simon Gosejohann being misrepresented by the model.
}

\subsection{Semantically weak first tokens}
\textsc{person\_university}
exemplifies the problem of 
\emph{semantically weak first tokens}. Many universities
have names that start with ``University''.
So the LM learns
to predict this common prefix as the target token for many
persons.
The standard setup is to count such a prediction as
correct. But this is not an effective evaluation:
The correct generation of ``University'' is not
a reliable indicator of the factual competence of the LM.

The collapse rate is the fraction of a relation's distinct
objects that share a first token with another. \textbf{6 of 47
relations exceed 25\%}, led by \textsc{person\_mother} (61\%),
\textsc{person\_father} (58\%) and \textsc{person\_university} (45\%:
29 of 69 objects begin with \textsc{University}).

\enote{hs}{cut for space
Restricting each relation to objects whose first token identifies them
repairs this on the gold side.
It is only a partial remedy: faithfulness (which
compares the LRE against the \emph{model's} top-1, not
gold) is
inflated by 
counting the model's own answer as ``faithful''. In
\textsc{person\_university}, after filtering, the model still answers
\textsc{University} for 7 of 29 subjects and the LRE agrees on every one.
A complete fix would condition on the prediction being distinctive.
The direction of the bias differs by relation. Where the collapse is on
the gold side it makes faithfulness easier: 41 distinct fathers all
count as \textsc{John}.
Where the model itself emits the shared token, the collapse manufactures
agreement that is not an accurate response at all.
}


Both LRE and the lens work on the token level, the residual
stream vector at the last layer, which is focused on
producing the correct next token and may not represent
information about the tokens to follow. This
single-token regime is problematic for the standard
setup because we are effectively investigating a relation in
which all objects have been shortened to their first
token. This 
is clearly different from the intended relations
whose objects are full entity names.

\subsection{Towards better fact datasets}
Our analysis motivates two requirements for future evaluations. (i)
Answer scoring should distinguish complete objects rather than credit a 
semantically weak first token.
\enote{hs}{restore if there is more space
Filtering for identifying first tokens provides
one diagnostic, but excludes some relations and does not ensure that the model's
own prediction is distinctive.}
Evaluation methods that score complete answers
are therefore needed for relations with unavoidable prefix overlap.
(ii) A subject name should refer to the intended entity for the model under
study. Dataset construction alone may not ensure this, because different models
can resolve the same name differently. Evaluations should examine the entity
that a model associates with each subject, document ambiguity, and distinguish
errors of subject identification from errors in relational decoding. These are
methodological directions motivated by the observed failures, rather than
solutions evaluated in this study.


\section{Limitations}
\label{sec:limitations}

The study only covers 28 relations,\enote{sg}{I think we should consistently say "relations", not "relationships" (for things like country-capital).}
some of which have small sample sizes.
Figure~\ref{fig:main}
is
correlational: we show that a relation-agnostic readout predicts LRE
faithfulness, not that the representation it reads is what the LRE
exploits.

Our NEL-based score is only a good predictor of LRE faithfulness in the middle of the object cardinality range
(\S\ref{sec:residuals}).
It is weakest where the answer space is very
small, since an LRE can then be faithful by elimination rather than by
reading anything from the subject, and where the answer space is very
large and each answer near-unique, since faithfulness is then close to its
floor regardless. We show that these are the two tails of a single
variable, but we do not have a principled description of where the boundaries
lie.


We also probe salience through a limited set of neutral prompts. That phrasing asks what an entity is \emph{noted} for, which need not coincide with everything its
representation carries; attributes such as gender may be well encoded
without being something a name is ``known for''. A different neutral
prompt, or an average over several, might track salience more broadly.

The smallest relations have only 19--24 examples, so their faithfulness
estimates remain coarse.
\enote{sg}{Removed sentence: "a
rotating-split protocol would sharpen these."
Whatever this means and however right this may be, I don't think this is helpful in a limitations section.
}

Finally, the diagnostic is
defined only for relations
whose subjects are entities, not for relations with
non-entity subjects.

\section{Conclusion}

Whether a relation is linearly decodable from an LM  is
substantially predictable without the relation. A Jacobian lens estimated
from subjects
\enote{sg}{changed "subject differences" to "subjects"}
under a neutral prompt, at a fixed layer, ranks
the model's own answer highly for exactly those relations whose LREs are
faithful.

We favor salience as the mechanism: LRE faithfulness depends on how readily
the subject's representation evokes the answer. On this view, variation
across relations largely reflects the entities they concern.


\enote{hs}{do we explain how the subset of facts was
  selected that the experiments are done for? for example, i
  think we're only using 100 from person-occupration? if
  not: add an appendix section on this and refer to it in
  the text}

\label{sec:main-end}
\clearpage

\ifdefined\arxivversion
\section*{Acknowledgements}
This work was supported by the German Research Foundation (DFG) under
grants SCHU~2246/14-1 and GU~798/29-1 (UKP-SQuARE project). 
This work was additionally supported by the DAAD programme Konrad Zuse Schools of Excellence in Artificial Intelligence, sponsored by the Federal Ministry of Research, Technology and Space.
We gratefully acknowledge support from Google through a generous GCP research grant.

\fi

\section*{AI use statement}
We used generative AI tools for brainstorming, assisting with software code, and initial drafting and rewriting of parts of the manuscript. All AI-assisted code and text were reviewed and verified by the authors, and generated code was tested for correctness. The authors take responsibility for the final content of this work, including all text, claims, results, and artifacts produced with the aid of generative AI.

\section*{Reproducibility statement}
 We release the complete pipeline, and one command reproduces every number, table and figure in the paper from scratch: it downloads the LRE dataset and all models, runs every measurement, and builds the tables and figures. To confirm this, we ran the full pipeline in a fresh Docker container and compared every reported value against the paper. The pipeline pins the datasets, checkpoints, prompt templates, random seeds and hyperparameters in configuration files. Reproducing everything takes roughly 35 A100 (80 GB) GPU-hours for the seven models, plus about 1,200 CPU core-hours and 3.6 TB of disk for the OLMo training-corpus counts. All models and datasets are publicly available.
\bibliographystyle{iclr2027_conference}
\bibliography{custom}

\clearpage
\appendix

\section{Baseline comparisons}
\label{sec:shared-baselines}
The first two tables report Spearman correlations with LRE faithfulness across the LRE Dataset.

\begin{table}[!htbp]
\centering
\caption{\textbf{Entity lenses and readout baselines across seven models.} Spearman correlation $\rho$ with LRE faithfulness.}
\label{tab:shared-lenses-spearman}
\begingroup
\small
\renewcommand{\arraystretch}{1.18}
\setlength{\tabcolsep}{3pt}
\begin{tabular*}{\linewidth}{@{\extracolsep{\fill}}lrrrrr@{}}
\toprule
Model & \shortstack[r]{Contrastive\\entity lens} & \shortstack[r]{Noncontrastive\\entity lens} & Logit lens & \shortstack[r]{Differential\\logit lens} & \shortstack[r]{Attribute\\rate} \\
\midrule
OLMo 3 7B & $\phantom{-}0.7355$ & $\phantom{-}\mathbf{0.7650}$ & $\phantom{-}0.2056$ & $\phantom{-}0.4390$ & $-0.4016$ \\
GPT-J 6B$^{*}$ & $\phantom{-}0.4346$ & $\phantom{-}0.4196$ & $\phantom{-}\mathbf{0.4669}$ & $\phantom{-}0.2020$ & $-0.1035$ \\
Mistral 7B & $\phantom{-}0.6561$ & $\phantom{-}\mathbf{0.6711}$ & $\phantom{-}0.6581$ & $\phantom{-}0.5186$ & $-0.2202$ \\
Gemma 2 9B & $\phantom{-}\mathbf{0.6889}$ & $\phantom{-}0.6868$ & $\phantom{-}0.4728$ & $\phantom{-}0.3436$ & $-0.5515$ \\
Qwen3 8B Base$^{*}$ & $\phantom{-}0.6647$ & $\phantom{-}0.6232$ & $\phantom{-}\mathbf{0.7908}$ & $\phantom{-}0.5952$ & $-0.0557$ \\
Qwen3.8 27B$^{*}$ & $\phantom{-}0.5539$ & $\phantom{-}\mathbf{0.5717}$ & $\phantom{-}0.3404$ & $\phantom{-}0.4492$ & $-0.0099$ \\
Pythia 6.9B$^{*,\dagger}$ & $\phantom{-}0.5027$ & $\phantom{-}\mathbf{0.5063}$ & $-0.2047$ & $\phantom{-}0.1845$ & $\phantom{-}0.0041$ \\
\bottomrule
\end{tabular*}
\endgroup
\par\vspace{5pt}
\begin{minipage}{\linewidth}\footnotesize
Bold marks the largest point estimate within each model, not a significant difference. All columns use the contrastive entity lens's same evaluation rows, including demonstrations and references, and the same model-specific year exclusions. $^{*}$GPT-J, Pythia, Qwen3 8B Base, and Qwen3.8 27B use one leading newline in the neutral prompt (\texttt{\textbackslash n\{s\} is known for}) to reduce excessive activation norms. The other models use the prompt without this prefix. $^{\dagger}$Pythia has 2,663 empty filtered prediction sets; Gemma has one. Their attribute rates are assigned zero, retaining every evaluation row.
\end{minipage}
\end{table}

\paragraph{Readout definitions.}
For subject state $s_e$, reference states $s_i$, and reference Jacobians
$J_i$, the contrastive and noncontrastive entity lenses decode
$\operatorname{mean}_i J_i(s_e-s_i)$ and
$\operatorname{mean}_i J_i s_e$, respectively. Logit lens decodes $s_e$;
differential logit lens decodes $s_e-\operatorname{mean}_i s_i$. Each relation's score is the negative median $\log_{10}(\mathrm{rank}+1)$ of the model's answer token.

\paragraph{Attribute rate.}
Following \citet{geva-etal-2023-dissecting}, we measure overlap between filtered top-50
logit-lens predictions and subject attributes retrieved from Wikipedia.
The index uses the union of current test subjects; the source stopword,
length, and retrieval filters are retained. Relation scores average the
percentage overlap. Undefined rates are handled as noted in the table;
matched-row sensitivity results remain in the accompanying data.

NEL has a larger point estimate than CEL in four models. All seven paired 95\% relation-bootstrap intervals for their
difference include zero; these results do not establish a general advantage
for either readout. LRE hyperparameter selection remains in sample.

\clearpage
\subsection{OLMo corpus-frequency baselines}
\label{sec:corpus-freq-baselines}

\begin{table}[!htbp]
\centering
\caption{\textbf{Frequency baselines for OLMo 3 7B.} Correlations with LRE faithfulness using document counts from OLMo's three published training mixes combined.}
\label{tab:shared-corpus-spearman}
\begingroup
\small
\renewcommand{\arraystretch}{1.16}
\setlength{\tabcolsep}{6pt}
\begin{tabular*}{\linewidth}{@{\extracolsep{\fill}}lrr@{}}
\toprule
Frequency measure & Spearman $\rho$ & $p$-value \\
\midrule
\multicolumn{3}{@{}l}{\textit{Marginal frequencies}} \\
\addlinespace[2pt]
Subject frequency, $c_s$ & $\phantom{-}0.1380$ & $0.484$ \\
Object frequency, $c_o$ & $\phantom{-}0.5712$ & $0.001$ \\
\addlinespace[5pt]
\multicolumn{3}{@{}l}{\textit{Co-occurrence}} \\
\addlinespace[2pt]
Joint frequency, $c_{so}$ & $\phantom{-}0.4568$ & $0.015$ \\
Mean of $\log(1+c_{so})$ & $\phantom{-}0.4191$ & $0.026$ \\
$\log(1+\mathrm{mean}(c_{so}))$ & $\phantom{-}0.4568$ & $0.015$ \\
\addlinespace[5pt]
\multicolumn{3}{@{}l}{\textit{Conditional frequencies}} \\
\addlinespace[2pt]
Forward conditional, $P(o\mid s)$ & $\phantom{-}0.4486$ & $0.017$ \\
Reverse conditional, $P(s\mid o)$ & $-0.4149$ & $0.028$ \\
Directional difference, $P(o\mid s)-P(s\mid o)$ & $\phantom{-}0.4826$ & $0.009$ \\
Micro-averaged $P(o\mid s)$ & $\phantom{-}0.5504$ & $0.002$ \\
Micro-averaged $P(s\mid o)$ & $-0.1453$ & $0.461$ \\
\bottomrule
\end{tabular*}
\endgroup
\par\vspace{5pt}
\begin{minipage}{\linewidth}\footnotesize
Counts denote binary document presence. Means use the current test rows with both marginals observed; micro averages are ratios of summed counts. The $p$-values are two-sided and unadjusted.
\end{minipage}
\end{table}

\paragraph{Counting and aggregation.}
A document contributes at most one mention per subject/object string.
Matching uses normalized literal names with Unicode word boundaries,
without alias resolution or a proximity constraint. With subject, object,
and joint counts $c_s$, $c_o$, and $c_{so}$, the conditionals are
$P(o\mid s)=c_{so}/c_s$ and $P(s\mid o)=c_{so}/c_o$.
Macro features average row-level quantities where both marginals are
observed. Micro conditionals divide summed joint counts by summed
marginals. The log of the mean joint count has the same Spearman correlation
as the mean joint count because the transformation is monotonic.

\paragraph{Interpretation and reproducibility.}
These are exploratory associations. Relations can share subjects and
objects, and the reported $p$-values are unadjusted. Conditional direction
also reflects marginal frequency: when $c_{so}>0$, $P(o\mid s)/P(s\mid o)=c_o/c_s$.

\clearpage
\subsection{Neutral-prompt variations}
\label{sec:prompt-variations}

Table~\ref{tab:shared-prompt-variations} reports prompt variations for the
Qwen3 8B Base contrastive entity lens. We tested these variations because
initial correlations were low with the neutral prompt used for OLMo,
\texttt{\{s\} is known for}. All prompts use the same checkpoint,
layer 19, demonstrations, reference sets, 9,747 evaluation rows, gender scoring,
relation-prompt answer tokens, and LRE outcomes as
Table~\ref{tab:shared-lenses-spearman}. Correlations use the same 26 relations
after the model's two year-relation exclusions. Only the displayed neutral
prompt varies; \texttt{\textbackslash n} denotes one leading newline.

After observing unusually large subject-state norms at position zero, we
added a leading newline (\texttt{\textbackslash n}) to the original prompt.
This moves every subject past position zero while preserving Qwen's original
token sequence. The table marks this control with $\dagger$.

\begin{table}[!htbp]
\centering
\caption{\textbf{Neutral-prompt variations.} Spearman $\rho$ between contrastive entity lens scores and LRE faithfulness, using the evaluation setup of Table~\ref{tab:shared-lenses-spearman}. $\dagger$: leading newline added to address high subject-state norms at position zero.}
\label{tab:shared-prompt-variations}
\begingroup
\small
\renewcommand{\arraystretch}{1.16}
\setlength{\tabcolsep}{5pt}
\begin{tabular*}{\linewidth}{@{\extracolsep{\fill}}lr@{}}
\toprule
\multicolumn{2}{@{}l}{\textbf{Qwen3 8B Base: prompt wording and prefix}} \\
\addlinespace[3pt]
Neutral prompt & $\rho$ \\
\midrule
\texttt{\{s\} is known for} & $\phantom{-}0.3430$ \\
\texttt{\{s\}} & $\phantom{-}0.2479$ \\
\texttt{\{s\} is} & $\phantom{-}0.2362$ \\
\texttt{\{s\} can be described as} & $\phantom{-}0.4297$ \\
\texttt{Information about \{s\}:} & $\phantom{-}0.6998$ \\
\texttt{Facts about \{s\}:} & $\phantom{-}0.6725$ \\
\texttt{Entity: \{s\} is known for} & $\phantom{-}0.7331$ \\
\addlinespace[4pt]
\texttt{\textbackslash n\{s\} is known for}$^{\dagger}$ & $\phantom{-}0.6647$ \\
\bottomrule
\end{tabular*}
\endgroup
\end{table}

\clearpage

\section{Jacobian--vector products}
\label{sec:jvp}

Both the operator and the lens access $J_i$ only through
Jacobian--vector products $J_i u$. This is explicit for the lens in
Eq.~\ref{eq:cel}; for the LRE, the expansion in Eq.~\ref{eq:beta}
expresses $W = \operatorname{mean}_i J_i$ as a sum of these products. We
compute each product using the standard double-backward construction
\citep{pearlmutter1994fast}: differentiate $\langle u', f(h)\rangle$
with respect to $h$, treating $u'$ as free, and then differentiate with
respect to $u'$ in direction $u$. This takes a constant number of forward
and backward passes, while materializing the Jacobian takes
$d_{\text{model}}$ passes. We use \texttt{float32} and eager attention
because fused attention kernels do not support second derivatives.

\section{Layer selection}
\label{sec:layer17}
Layer 17 is roughly half-way through the network,
the region in which such lenses are usable at all
\citep{gurnee2026workspace}, and is the modal best layer for the LRE
itself in our sweep (15 of 28 relations; the rest spread over 5, 11, 14,
20 and 23). A single layer is used for \emph{every}
relation, without per-relation tuning. However, this does not establish robustness to
the layer choice: changing the layer can affect relations differently.
A systematic sensitivity analysis is left to future work.


%

\section{Numeric-object tokenization correction}
\label{sec:numeric-objects}

Both relations with numeric objects
(\textsc{presidents\_birth\_year}, \textsc{presidents\_election\_year})
initially scored faithfulness $1.000$. This was spurious. In this
tokenizer \texttt{" 1743"} segments as \texttt{[\,' ', '174', '3'\,]}:
digits never merge with a preceding space. First-token scoring therefore
compared the model's top-1 against a bare space token on both sides, and
faithfulness $1.000$ meant only that the LRE and the model both emit a
space.

The fix is to place the space in the prompt, so that the query ends with
it and the next token is content. Rescored, the two relations give
faithfulness $0.74$ and $0.79$ and lens scores of $-4.44$ and $-3.45$; the model does
predict the year correctly once the space is consumed. Including the
uncorrected values would have depressed the reported correlation from
$\headlinenumber$ to $0.416$.

\section{Recombining anchors and subject terms}
\label{sec:swapdetail}

This appendix describes the experiments of \S\ref{sec:swap} in full.

\paragraph{The decomposition.} We write the LRE in the form of
Eq.~\ref{eq:basiclre}, $\beta W s + b_\beta$, and call $b_\beta$ the
\emph{anchor} and $\beta W s = \beta \operatorname{mean}_i J_i s$ the
\emph{subject term}. The anchor is fixed by the reference subjects
alone; the subject term is NEL (Eq.~\ref{eq:nel}) with the Jacobians
taken under the relation's prompt. Their sum is the operator exactly, so
the \emph{own} arm is the unmodified LRE. The \emph{foreign} arm replaces
$\beta W_r s$ by $\beta W_{r'} s$, the paired relation's subject term,
and keeps $b_\beta^r$. Because the constant
$\operatorname{mean}_i J_i s_i$ stays inside the anchor $b_\beta$, a
foreign subject term imports the other relation's $W$ and nothing else.
The \emph{NEL} arm replaces the subject term by $\beta\,\mbox{NEL}(e)$
under the neutral prompt; \emph{anchor alone} is $b_\beta$; the
\emph{wrong donor} is $b_\beta + \beta\,\mbox{NEL}(e')$ for a different
subject $e'$.

\paragraph{Choice of relation pairs.} A term can be swapped with nothing
else changing only if the two relations share their subjects, and a
substitute can only be measured against an operator that produces the
answer in the first place. In the benchmark two pairs meet both
conditions: \textsc{country\_capital\_city} with
\textsc{country\_largest\_city}, and
\textsc{person\_plays\_position\_in\_sport} with
\textsc{person\_plays\_pro\_sport}, over the 24 athletes the two share.
Two further pairs fail the second condition.
\textsc{country\_currency} shares its countries with the other country
relations, but its own operator recovers the currency for only 1 of 10
countries at $\beta = 3$, 0 at $\beta = 5$ and none at $\beta = 10$;
where an operator is this weak there is nothing for a substitute to
reproduce. \textsc{person\_father} and \textsc{person\_mother} would be
the strongest test: they share 748 subjects and have answers of the same
kind, so the two relations cannot be told apart by the kind of answer.
But their faithfulness is $0.07$ and $0.07$, and the operators essentially
never produce the gold object.

\paragraph{Country subjects.} A country tests the swap only if its
capital differs from its largest city; otherwise one answer satisfies
both relations and a hit cannot say which relation won. The benchmark
has 9 such countries among the 18 that the two relations share. We
extended the set from the list of sovereign states whose capital is not
their largest city.\footnote{\url{https://en.wikipedia.org/wiki/List_of_countries_whose_capital_is_not_their_largest_city},
retrieved 2026-09-24.} Of its 37 states we excluded seven whose capital
or largest city the list marks as contested: Israel and Palestine
(limited recognition), Taiwan (sovereignty), Nauru (de facto capital
only), Bolivia (La Paz is the seat of government), South Africa (three
capitals) and Equatorial Guinea (capital in transition). Of the 31
remaining we kept those the model answers correctly for \emph{both}
relations under the prompts used here, scoring the model's greedy
continuation rather than its first token, with diacritics folded. All 9
benchmark countries and 13 of the 22 added ones pass, giving 22:
Australia, Belize, Benin, Brazil, Cameroon, Canada, China, India, Ivory
Coast, Kazakhstan, Morocco, Myanmar, New Zealand, Nigeria, Pakistan,
Palau, Switzerland, Tanzania, Turkey, the United Arab Emirates, the
United States and Vietnam. Every rejected country is one for which the
model gives the \emph{same} city for both questions: the capital for
the largest city (Ecuador, Gambia, Liechtenstein, Malta, Micronesia, the
Philippines, San Marino, Trinidad and Tobago), or the largest city for
the capital (Burundi). The filter therefore removes exactly the countries
on which the model does not distinguish the two relations.

\paragraph{Protocol.} All states are read at layer 17. Each relation's
prompt is its first template with eight in-context demonstrations drawn
from its own samples outside the candidate set, so no test subject
appears in any prompt. Each test subject is read with the remaining test
subjects as references, for the operators and for NEL alike. We fix
$\beta = 3$. It was chosen by inspecting an earlier version of this
analysis on the same 24 athletes and on 12 countries, 7 of which are
among the 22, so it is not independent of the evaluation; it is not
re-tuned here. At this value the subject term is
smaller than the anchor, $\beta\lVert W s\rVert / \lVert b_\beta\rVert
\approx 0.8$ for the country relations and $0.4$--$0.6$ for the person
relations. A hit is a decoded top-1 equal to the first token of the gold
object; since the model answers every test country correctly, gold and
model answer coincide for the country pair.

\enote{claude}{Sensitivity to $\beta$ has not yet been re-run for this
form of the decomposition.}

\paragraph{Full results.} Table~\ref{tab:swapfull} adds one arm to
Table~\ref{tab:swap}: a subject term estimated under the prompt
\texttt{"The official currency of \{\} is the"}, which shares nothing
with either country relation. Conditional on the operator being right,
the foreign subject term preserves the answer for 11 of 12 capitals and
8 of 12 largest cities, NEL for 7 of 12 and 4 of 12, and the currency
subject term for 6 of 12 and 5 of 12.

\begin{table}[h]
\centering
\small
\begin{tabular}{lrrrrrrr}
\toprule
relation & $n$ & own & foreign & NEL & currency & anchor alone & wrong donor \\
\midrule
\textsc{country\_capital\_city} & 22 & 12 & 12 & 7 & 6 & 0 & 0 \\
\textsc{country\_largest\_city} & 22 & 12 & 8 & 4 & 5 & 0 & 0 \\
\textsc{person\_plays\_position} & 24 & 19 & 21 & 18 & -- & 7 & 6 \\
\textsc{person\_plays\_pro\_sport} & 24 & 22 & 21 & 21 & -- & 6 & 3 \\
\bottomrule
\end{tabular}
\caption{All arms, $\beta = 3$, each subject read with the others as
references. \emph{Currency}: a subject term estimated under the
official-currency prompt, which shares nothing with either country
relation.}
\label{tab:swapfull}
\end{table}

\paragraph{Where the operator fails.} Although the model answers all 22
countries correctly, the operator reproduces only 12 per relation. It
does much better on the benchmark's nine (7 capitals, 7 largest cities)
than on the 13 added countries (5 and 5), which are rarer in text. On
those it fails in recognizable ways: it echoes the country's own name
(Kazakhstan, Tanzania, Vietnam), names a famous associated city (Paris
for Ivory Coast and Morocco, Geneva for Switzerland), or produces a
degenerate token. This is consistent with linear structure weakening for
entities that are rarer in pretraining \citep{merullo2025linear}.

\paragraph{Dependence on the reference sets.} In the leave-one-out
design above the references are the other test subjects, so they barely
vary. To measure how much the results depend on them we held every
subject term fixed and drew the \emph{anchor's} eight references at
random, 500 times, from subjects outside the test set and outside every
demonstration: 16 countries, and a fixed random 150 of the 1{,}206
eligible athletes in the two sport relations. Table~\ref{tab:swapanchor} gives
the result. The operator and the foreign subject term barely move; NEL
is the arm that depends on the anchor, most for position, whose worst
anchor brings it down to the anchor-alone level.

\begin{table}[h]
\centering
\small
\begin{tabular}{lrrrr}
\toprule
arm & capital (22) & largest (22) & position (24) & sport (24) \\
\midrule
own & 10.7 (10--14) & 11.3 (10--12) & 18.3 (17--20) & 22.0 (21--23) \\
foreign & 9.7 (9--11) & 9.3 (8--10) & 17.3 (15--21) & 20.4 (16--23) \\
NEL & 5.2 (1--8) & 4.6 (4--5) & 14.7 (7--20) & 21.5 (16--24) \\
anchor alone & 0.0 (0--1) & 0.0 (0--0) & 7.0 (7--7) & 4.5 (1--7) \\
\bottomrule
\end{tabular}
\caption{Mean (range) hits over 500 anchors, each estimated from eight
references drawn from outside the test set; subject terms fixed.}
\label{tab:swapanchor}
\end{table}

Part of the spread for position is composition. The athlete pool has
fewer baseball players and more hockey players than the test set, and an
anchor with more soccer players gives fewer position hits (Spearman
$\rho = -0.32$ across anchors), plausibly because the anchor's typical
position leans toward soccer's.

For the country pair we also varied the two reference sets jointly: 40
anchors crossed with 40 sets of references for NEL, drawn from the 30
countries outside the test set. Of the variance in NEL's hits, the
anchor's references account for $36\%$ (capital) and $25\%$ (largest
city), NEL's references for $38\%$ and $42\%$, and their interaction for
the rest. Neither source dominates.

\paragraph{Scoring.} Hits compare the first token of the decoded output
with the first token of the gold object. Three cases in the country pair
are affected: the model writes S\~{a}o Paulo as ``Sao'', a different
first token, and the gold ``New Delhi'' begins with ``New'', so the
answer ``Delhi'' scores as a miss; conversely the first token of
Ngerulmud, Palau's capital, is the single letter ``N'', so a bare ``N''
scores as a hit. In the own column these roughly cancel. Scoring the full
continuation would remove all three.

\section{All relations}

Table~\ref{tab:relations} gives faithfulness and the lens score for every
relation in the study. Table~\ref{tab:prominence} gives object prominence
and the lens's top-five hit rate for the 14 relations on which prominence
can be measured (Section~\ref{sec:residuals}).

\begin{table}[t]
\centering\small
\caption{All 28 relations, sorted by LRE faithfulness. $k$ is the mean number
of subjects per distinct object; $A_r$ is the relation-agnostic lens score, higher
meaning closer agreement with the model. $^{\dagger}$rescored for the
numeric-object artifact (Appendix~\ref{sec:numeric-objects}); $^{*}$fewer than ten test items,
so faithfulness is coarsely quantized.}
\label{tab:relations}
\begin{tabular}{@{}lrrr@{}}
\toprule
relation & $k$ & faith. & $A_r$ \\
\midrule
country\_language$^{*}$ & 1.7 & 1.00 & $-1.22$ \\
name\_birthplace & 3.9 & 0.97 & $-1.32$ \\
name\_religion & 6.2 & 0.97 & $-1.40$ \\
person\_plays\_pro\_sport & 63.6 & 0.95 & $-1.23$ \\
landmark\_on\_continent & 189.4 & 0.92 & $-2.43$ \\
country\_largest\_city$^{*}$ & 1.0 & 0.92 & $-0.81$ \\
person\_native\_language & 30.6 & 0.92 & $-1.69$ \\
person\_plays\_instrument & 85.5 & 0.89 & $-1.72$ \\
city\_in\_country & 1.3 & 0.89 & $-0.48$ \\
country\_capital\_city$^{*}$ & 1.0 & 0.88 & $-0.85$ \\
person\_band\_lead\_singer$^{*}$ & 1.0 & 0.86 & $-1.49$ \\
landmark\_in\_country & 9.2 & 0.80 & $-1.69$ \\
presidents\_election\_year$^{\dagger}$ & 1.1 & 0.79 & $-3.45$ \\
presidents\_birth\_year$^{\dagger}$ & 1.3 & 0.74 & $-4.44$ \\
person\_occupation & 26.5 & 0.71 & $-2.74$ \\
product\_by\_company & 17.4 & 0.68 & $-1.88$ \\
person\_plays\_position\_in\_sport & 68.0 & 0.61 & $-3.64$ \\
person\_university & 1.3 & 0.59 & $-4.38$ \\
company\_hq & 4.1 & 0.54 & $-4.23$ \\
name\_gender$^{*}$ & 9.5 & 0.53 & $-1.04$ \\
superhero\_person & 1.0 & 0.50 & $-2.62$ \\
star\_constellation & 11.7 & 0.47 & $-3.50$ \\
country\_currency & 1.3 & 0.40 & $-4.49$ \\
pokemon\_evolutions & 1.0 & 0.32 & $-3.45$ \\
superhero\_archnemesis & 1.1 & 0.30 & $-3.90$ \\
person\_father & 1.0 & 0.07 & $-3.71$ \\
person\_mother & 1.0 & 0.07 & $-4.18$ \\
company\_ceo & 1.0 & 0.05 & $-4.50$ \\
\bottomrule
\end{tabular}
\end{table}

\section{The anchored contrastive entity lens}
Here we report on an experiment with an ``anchored''
contrastive entity lens rather than the non-anchored
entity lens defined in Eq.~\ref{eq:cel}.
Keeping the anchor requires choosing how to weight the anchor and subject
term, which is what $\beta$ does in Eq.~\ref{eq:beta}.

In sample
no such $\beta$ outperforms dropping the anchor: the correlation rises
towards the anchor-free value as $\beta$ grows, and selection does not
help either: choosing $\beta$ on 15 relations and evaluating on the
remaining 13, across 2{,}000 random splits, beats the anchor-free readout
in only $1\%$ of them. We therefore drop the anchor, which leaves a
readout with no hyperparameter at all. The decoder normalizes, so once
the anchor is gone, scaling cannot change a ranking and $\beta$ has
nothing left to do. This gives the
\emph{contrastive entity lens} that we defined  in Eq.~\ref{eq:cel}.

In summary, a single
$\beta$ shared by every relation would be a legitimate option, adding
nothing relation-specific, but it turns out to have no
advantage compared to the simpler form we adopted.

\end{document}